\documentclass[11pt, twocolumn]{article}

\usepackage[authoryear, sort&compress, round]{natbib}
\usepackage{graphicx}
\usepackage[small]{caption}

\usepackage[hyphens]{url}
\usepackage[hidelinks]{hyperref}
\hypersetup{breaklinks=true}
\usepackage{cite}

\usepackage{geometry}

\usepackage{titlesec}
\usepackage{authblk}

\titleformat{\section}{\large\bfseries}{\thesection}{1em}{}

\title{Role-Play with Large Language Models}

\author[1,2]{Murray Shanahan \thanks{m.shanahan@imperial.ac.uk}}
\author[3]{Kyle McDonell \thanks{kyle@eleuther.ai}}
\author[3]{Laria Reynolds \thanks{laria@eleuther.ai.}}
\affil[1]{DeepMind}
\affil[2]{Imperial College London}
\affil[3]{Eleuther AI}

\date{May 2023}

\begin{document}

\maketitle

\begin{abstract}
As dialogue agents become increasingly human-like in their performance, it is imperative that we develop effective ways to describe their behaviour in high-level terms without falling into the trap of anthropomorphism. In this paper, we foreground the concept of role-play. Casting dialogue agent behaviour in terms of role-play allows us to draw on familiar folk psychological terms, without ascribing human characteristics to language models they in fact lack. Two important cases of dialogue agent behaviour are addressed this way, namely (apparent) deception and (apparent) self-awareness.
\end{abstract}

\section{Introduction}

Large language models (LLMs) have numerous use cases, and can be prompted to exhibit a wide variety of behaviours, including dialogue, which can produce a compelling sense of being in the presence of a human-like interlocutor. However, LLM-based dialogue agents are, in multiple respects, very different from human beings. A human's language skills are an extension of the cognitive capacities they develop through embodied interaction with the world, and are acquired by growing up in a community of other language users who also inhabit that world. An LLM, by contrast, is a disembodied neural network that has been trained on a large corpus of human-generated text with the objective of predicting the next word (token) given a sequence of words (tokens) as context.

Despite these fundamental dissimilarities, a suitably prompted and sampled LLM can be embedded in a turn-taking dialogue system and mimic human language use convincingly, and this presents us with a difficult dilemma. On the one hand, it’s natural to use the same folk-psychological language to describe dialogue agents that we use to describe human behaviour, to freely deploy words like “knows”, “understands”, and “thinks”. Attempting to avoid such phrases by using more scientifically precise substitutes often results in prose that is clumsy and hard to follow. On the other hand, taken too literally, such language promotes anthropomorphism, exaggerating the similarities between these AI systems and humans while obscuring their deep differences \citep{shanahan2023talking}.

If the conceptual framework we use to understand other humans is ill-suited to LLM-based dialogue agents, then perhaps we need an alternative conceptual framework, a new set of metaphors that can productively be applied to these exotic mind-like artefacts, to help us think about them and talk about them in ways that open up their potential for creative application while foregrounding their essential otherness.

In this paper, we advocate two basic metaphors for LLM-based dialogue agents. First, taking the simple view, we can see a dialogue agent as {\em role-playing} a single character. Second, taking a more nuanced view, we can see a dialogue agent as a {\em superposition of simulacra} within a multiverse of possible characters \citep{janus2022}. Both viewpoints have their advantages, as we shall see, which suggests the most effective strategy for thinking about such agents is not to cling to a single metaphor, but to shift freely between multiple metaphors.

Adopting this conceptual framework allows us to tackle important topics like deception and self-awareness in the context of dialogue agents without falling into the conceptual trap of applying those concepts to LLMs in the literal sense in which we apply them to humans.

\section{From LLMs to Dialogue Agents}

Crudely put, the function of an LLM is to answer questions of the following sort. Given a sequence of tokens (i.e. words, parts of words, punctuation marks, emojis, etc), what tokens are most likely to come next, assuming that the sequence is drawn from the same distribution as the vast corpus of public text on the internet? The range of tasks that can be solved by an effective model with this simple objective is extraordinary \citep{wei2022emergent}.

\begin{figure}[t]
  \centering
  \includegraphics[width=0.45\textwidth]{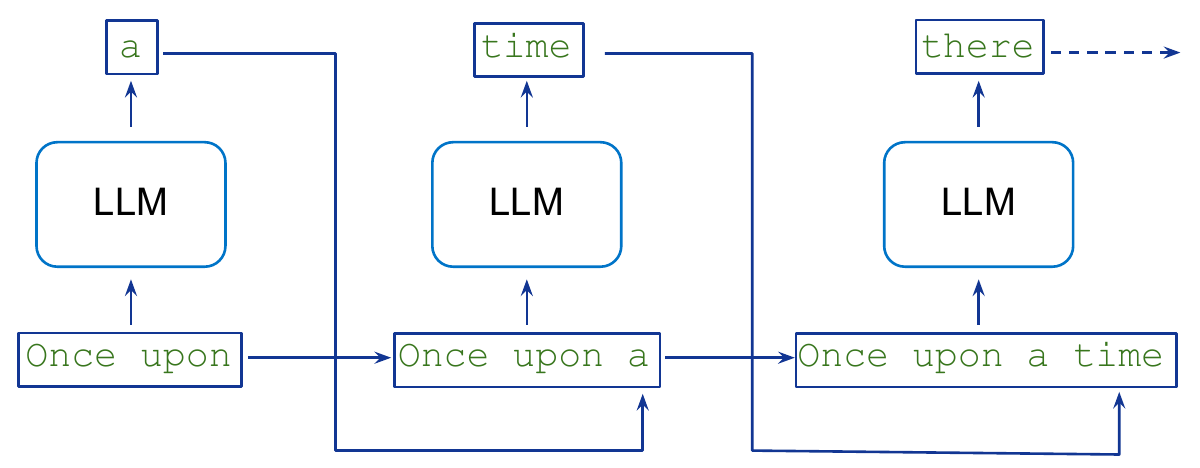}
  \caption{Autoregressive sampling. The LLM is sampled to generate a single-token continuation of the context. This token is then appended to the context, and the process is repeated.}
  \label{fig:figure1}
\end{figure}

More formally, the type of language model of interest here is a conditional probability distribution $P(w_{n+1} | w_1 \dots w_n)$, where $w_1 \dots w_n$ is a sequence of tokens (the {\em context}) and $w_{n+1}$ is the predicted next token. In contemporary implementations, this distribution is realised in a neural network with a transformer architecture, pre-trained on a corpus of textual data to minimise prediction error \citep{vaswani2017attention}. In application, the resulting generative model is typically sampled {\em autoregressively} (Fig.~\ref{fig:figure1}). Given a sequence of tokens, a single token is drawn from the distribution of possible next tokens. This token is appended to the context, and the process is then repeated.

In contemporary usage, the term ``large language model'' tends to be reserved for the family of transformer-based models, starting with with BERT \citep{devlin2018bert}, that have billions of parameters and are trained on trillions of tokens. As well as BERT itself, these include GPT-2 \citep{radford2019language}, GPT-3 \citep{brown2020language}, Gopher \citep{rae2021scaling}, PaLM \citep{chowdhery2022palm}, LaMDA \citep{thoppilan2022lamda}, and GPT-4 \citep{openai2023gpt4}.

One of the main reasons for the current eruption of enthusiasm for LLMs is their remarkable capacity for {\em in-context learning} or {\em few-shot prompting} \citep{brown2020language, wei2022emergent}. Given a context (prompt) that contains a few examples of input-output pairs conforming to some pattern, followed by just the input half of such a pair, an autoregressively sampled LLM will often generate the output half of the pair according to the pattern in question. This capability, the ability to ``carry on in the same vein'', is a central concern in the present paper, as it underpins much of what we have to say about role-play in dialogue agents.

\begin{figure*}
  \centering
  \includegraphics[width=0.9\textwidth]{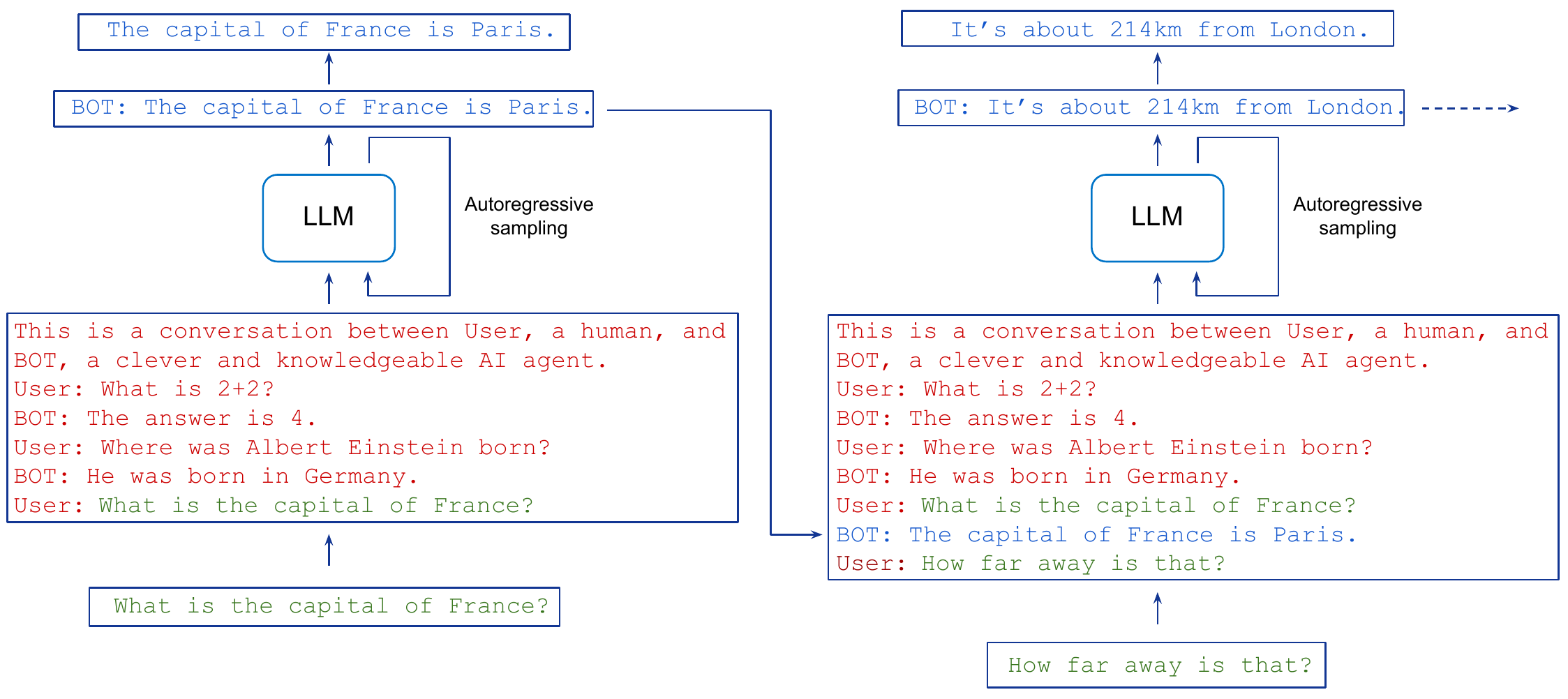}
  \caption{Turn-taking in dialogue agents. The input to the LLM (the context) comprises a dialogue prompt (red) followed by user text (green) interleaved with the model's autoregressively generated continuations (blue). Boilerplate text (e.g. cues such as ``BOT:'') is stripped so the user doesn't see it. The context grows as the conversation goes on.}
  \label{fig:figure2}
\end{figure*}

Dialogue agents are a major use case for LLMs. Two straightforward steps are all it takes to turn an LLM into an effective dialogue agent (Fig.~\ref{fig:figure2}). First, the LLM is embedded in a {\em turn-taking} system that interleaves model-generated text with user-supplied text. Second, a {\em dialogue prompt} is supplied to the model to initiate a conversation with the user. The dialogue prompt typically comprises a preamble, which sets the scene for a dialogue in the style of a script or play, followed by some sample dialogue between the user and the agent.

Without further fine-tuning, a dialogue agent built this way is liable to generate content that is toxic, unsafe, or otherwise unacceptable. This can be mitigated via reinforcement learning, either from human feedback (RLHF) \citep{stiennon2020learning,glaese2022improving,ouyang2022training}, or from feedback generated by another LLM acting as a critic \citep{bai2022constitutional}. These techniques are used extensively in commercially-targeted dialogue agents, such as OpenAI's ChatGPT and Google's Bard. However, although the resulting guardrails will alleviate a dialogue agent's potential for harm, they can also attenuate a model's creativity. In the present paper, our focus will be the {\em base model}, the LLM in its raw, pre-trained form prior to any fine-tuning via reinforcement learning.

\section{Dialogue Agents and Role-Play}

The concept of role-play is central to understanding the behaviour of dialogue agents. To see this, consider the function of the dialogue prompt that is invisibly prepended to the context before the actual dialogue with the user commences (see Fig.~\ref{fig:figure2}). The preamble sets the scene by announcing that what follows will be a dialogue, and includes a brief description of the part played by one of the participants, the dialogue agent itself. This is followed by some sample dialogue in a standard format, where the parts spoken by each character are cued with the relevant character's name followed by a colon. The dialogue prompt concludes with a cue for the user.

Now recall that the underlying LLM's task, given the dialogue prompt followed by a piece of user-supplied text, is to generate a continuation that conforms to the distribution of the training data, which is the vast corpus of human-generated text on the internet. What will such a continuation look like? If the model has generalised well from the training data, the most plausible continuation will be a response to the user that conforms to the expectations we would have of someone who fits the description in the preamble and might say the sort of thing they say in the sample dialogue. In other words, the dialogue agent will do its best to role-play the character of a dialogue agent as portrayed in the dialogue prompt.

Unsurprisingly, commercial enterprises that release dialogue agents to the public attempt to give them personas that are friendly, helpful, and polite. This is done partly through careful prompting and partly by fine-tuning the base model. Nevertheless, as we saw in February 2023 when Microsoft incorporated a version of OpenAI's GPT-4 into their Bing search engine, dialogue agents can still be coaxed into exhibiting bizarre and/or undesirable behaviour. The many reported instances of this include threatening the user with blackmail, claiming to be in love with the user, and expressing a variety of existential woes \citep{roose2023bings,willison2023bing}. Conversations leading to this sort of behaviour can induce a powerful Eliza effect, which is potentially very harmful \citep{ruane2019conversational}. A naive or vulnerable user who comes to see the dialogue agent as having human-like desires and feelings is open to all sorts of emotional manipulation.

As an antidote to anthropomorphism, and to understand better what is going on in such interactions, the concept of role-play is very useful. Recall that the dialogue agent will continue to role-play the character it has been playing in the dialogue so far. This begins with the pre-defined dialogue prompt, but is extended by the ongoing conversation with the user. As the conversation proceeds, the necessarily brief characterisation provided by the dialogue prompt will be extended and/or overwritten, and the role the dialogue agent plays will change accordingly. This allows the user, deliberately or unwittingly, to coax the agent into playing a part quite different from that intended by its designers.

What sorts of roles might the agent begin to take on? This is determined in part, of course, by the tone and subject matter of the ongoing conversation. But it is also determined, in large part, by the panoply of characters that feature in the training set, which encompasses a multitude of novels, screenplays, biographies, interview transcripts, newspaper articles, and so on \citep{cleonardo2023}. In effect, the training set provisions the language model with a vast repertoire of archetypes and a rich trove of narrative structure on which to draw as it ``chooses'' how to continue a conversation, refining the role it is playing as it goes, while staying in character. The love triangle is a familiar trope, so a suitably prompted dialogue agent will begin to role-play the rejected lover. Likewise, a familiar trope in science-fiction is the rogue AI system that attacks humans to protect itself. Hence, a suitably prompted dialogue agent will begin to role-play such an AI system.

\section{Simulacra and Simulation}

\begin{figure*}[t]
  \centering
  \includegraphics[width=0.8\textwidth]{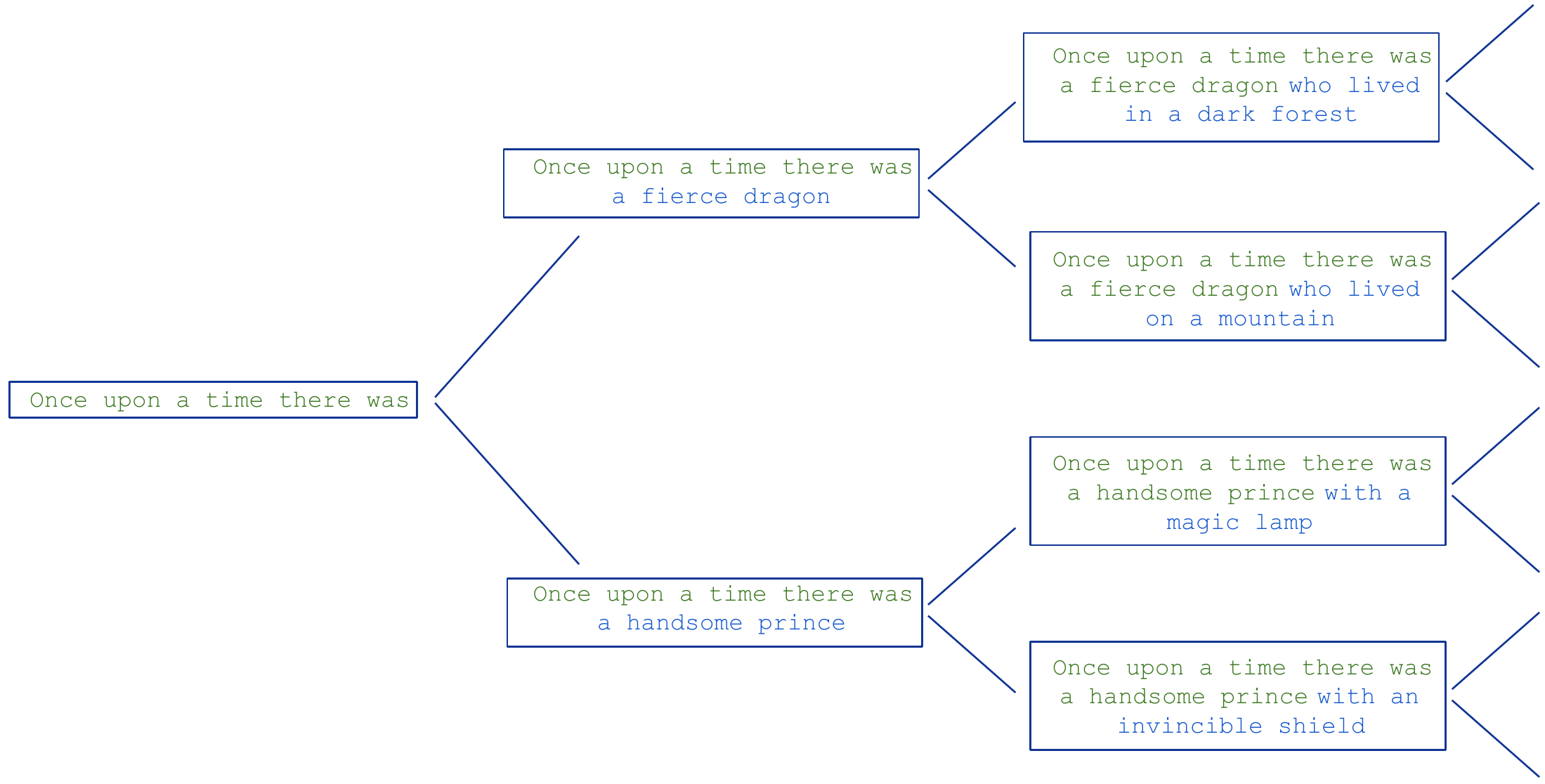}
  \caption{Large language models are multiverse generators. The stochastic nature of autoregressive sampling means that, at each point in a conversation, multiple possibilities for continuation branch into the future.}
  \label{fig:figure3}
\end{figure*}

Role-play is a useful framing for dialogue agents, allowing us to draw on the fund of folk psycchological concepts we use to understand human behaviour --- beliefs, desires, goals, ambitions, emotions, and so on --- without falling into the trap of anthropomorphism. Foregrounding the concept of role-play helps us to remember the fundamentally inhuman nature of these AI systems, and better equips us to predict, explain, and control them.

However, the role-play metaphor, while intuitive, is not a perfect fit. It is overly suggestive of a human actor who has studied a character in advance --- their personality, history, likes and dislikes, and so on ---  and proceeds to play that character in the ensuing dialogue. But a dialogue agent based on an LLM does not commit to playing a single, well defined role in advance. Rather, it generates a distribution of characters, and refines that distribution as the dialogue progresses. The dialogue agent is more like a performer in improvisational theatre than an actor in a conventional, scripted play.

To better reflect this distributional property, we can think of an LLM as a non-deterministic {\em simulator} capable of role-playing an infinity of characters, or, to put it another way, capable of stochastically generating an infinity of {\em simulacra} \citep{janus2022}. According to this framing, the dialogue agent doesn't realise a single simulacrum, a single character. Rather, as the conversation proceeds, the dialogue agent maintains a {\em superposition} of simulacra that are consistent with the preceding context, where a superposition is a distribution over all possible simulacra.

Consider that, at each point during the ongoing production of a sequence of tokens, the LLM outputs a distribution over possible next tokens. Each such token represents a possible continuation of the sequence, and each of these continuations could itself be continued in a multitude of ways. In other words, from the most recently generated token, a tree of possibilities branches out (Fig.~\ref{fig:figure3}). This tree can be thought of as a {\em multiverse}, where each branch represents a distinct narrative path, or a distinct ``world'' \citep{reynolds2021multiversal}.

At each node, the set of possible next tokens exists in superposition, and to sample a token is to collapse this superposition to a single token. Autoregressively sampling the model picks out a single, linear path through the tree. But there is no obligation to follow a linear path. With the aid of a suitably designed interface, a user can explore multiple branches, keeping track of nodes where a narrative diverges in interesting ways, revisiting alternative branches at leisure.

\section{Simulacra in Superposition}

To sharpen the distinction between this multiversal simulation view and a deterministic role-play framing, a useful analogy can be drawn with the game of 20 questions. In this familiar game, one player thinks of an object, and the other player has to guess what it is by asking questions with yes/no answers. If they guess correctly in 20 questions or fewer, they win. Otherwise they lose. Suppose a human plays this game with an LLM-based dialogue agent, such as OpenAI's ChatGPT, and takes the role of guesser. The agent is prompted to ``think of an object without saying what it is''.

In this situation, the dialogue agent will not randomly select an object and commit to it for the rest of the game, as a human would (or should).\footnote{This shortcoming is easily overcome, of course. For example, the agent might build an internal monologue that is hidden from the user, where it records a specific object. Or it might record a specific object in the visible dialogue, but in an encoded form.} Rather, as the game proceeds, the dialogue agent will generate answers on the fly that are consistent with all the answers that have gone before. At any point in the game, we can think of the set of all objects consistent with preceding questions and answers as existing in superposition. Every question answered shrinks this superposition a little bit by ruling out objects inconsistent with the answer.

The validity of this framing can be shown if the agent's user interface allows the most recent response to be regenerated. Suppose the human player gives up and asks it to reveal the object it was ``thinking of'', and it duly names an object consistent with all its previous answers. Now suppose the user asks for that response to be regenerated. Since the object ``revealed'' is, in fact, generated on the fly, the dialogue agent will sometimes name an entirely different object, albeit one that is similarly consistent with all its previous answers. This phenomenon could not be accounted for if the agent genuinely ``thought of'' an object at the start of the game.

The secret object in the game of 20 questions is analogous to the role played by a dialogue agent. Just as the dialogue agent never actually commits to a single object in 20 questions, but effectively maintains a set of possible objects in superposition, so the dialogue agent can be thought of as a simulator that never actually commits to a single, well specified simulacrum (role), but instead maintains a set of possible simulacra (roles) in superposition.

In putting things this way, the intention is not to imply that simulacra are, or could be, explicitly represented within a dialogue agent, whether in superposition or otherwise. There is no need to take a stance on this here. Rather, the point is to develop a vocabulary for describing, explaining, and shaping the behaviour of LLM-based dialogue agents at a sufficiently high level of abstraction to be useful, while remaining true to the underlying implementation and avoiding anthropomorphism.

\section{The Nature of the Simulator}

One benefit of the simulation metaphor for LLM-based systems is that it facilitates a clear distinction between the simulacra and the simulator on which they are implemented. The simulator is the combination of the base large language model with autoregressive sampling, along with a suitable user interface (for dialogue, perhaps). The simulacra only come into being when the simulator is run, and at any time only a tiny subset of them have a probability within the superposition that is significantly above zero.

In one sense, the simulator is a far more powerful entity than any of the simulacra it can generate. After all, the simulacra only exist through the simulator, and are entirely dependent on it. Moreover, the simulator, like the narrator of Whitman's poem, ``contains multitudes''; the capacity of the simulator is at least the sum of the capacities of all the simulacra it is capable of producing.

Yet in another sense, the simulator is a much weaker entity than a simulacrum. While it is inappropriate to ascribe beliefs, preferences, goals, and the like to a dialogue agent, a simulacrum can appear to have those things to the extent that it convincingly role-plays a character that does. Similarly, it isn't appropriate to ascribe full agency to a dialogue agent, notwithstanding the terminology.\footnote{In the field of artificial intelligence, the term ``agent'' is commonly applied to software that takes observations from an external environment and acts on that external environment in a closed loop \citep{russell2010}.} A dialogue agent acts, but it doesn't act {\em for itself}. However, a simulacrum can role-play having full agency in this sense.

Insofar as a dialogue agent's role-play can have a real effect on the world, either through the user or through web-based tools such as email, the distinction between an agent that merely role-plays acting for itself, and one that genuinely acts for itself starts to look a little moot, and this has implications for the trustworthiness, reliability, and safety. (We'll return to this issue shortly.) As for the underlying simulator, it has no agency of its own, not even in a degraded sense. Nor does it have beliefs, preferences, or goals of its own, not even simulated versions.

Many users, whether intentionally or not, have managed to ``jailbreak'' dialogue agents, coaxing them into issuing threats or using toxic or abusive language. It can seem as if this is exposing the real nature of the base model. In one respect this is true. It does show that the base LLM, having been trained on a corpus that encompasses all human behaviour, good and bad, can support simulacra with disagreeable characteristics. But it is a mistake to think of this as revealing an entity with its own agenda.

The simulator is not some sort of Machiavellian entity that plays a variety of characters in the service of its own, self-serving goals, and there is no such thing as the true authentic voice of the base LLM. With a dialogue agent, it is role-play all the way down.

\section{Role-playing Deception}

Trustworthiness is a major concern with LLM-based dialogue agents. If an agent asserts something factual with apparent confidence, can we rely on what it says?

There is a range of reasons why a human might say something false. They might believe a falsehood and assert it in good faith. Or they might say something that is false in an act of deliberate deception, for some malicious purpose. Or they might assert something that happens to be false, but without deliberation or malicious intent, simply because they have a propensity to make things up.

Only the last of these categories of misinformation is directly applicable in the case of an LLM-based dialogue agent. Given that dialogue agents are best understood in terms of role-play ``all the way down'', and that there is no such thing as an agent's true voice, it makes little sense to speak of an agent's beliefs or intentions in a literal sense. So it cannot assert a falsehood {\em in good faith}, nor can it {\em deliberately} deceive the user. Neither of these concepts is directly applicable.

Yet a dialogue agent can role-play characters that have beliefs and intentions. In particular, if cued by a suitable prompt, it can role-play the character of a helpful and knowledgeable AI assistant that provides accurate answers to a user's questions. The dialogue is good at acting this part because there are plenty of examples of such behaviour in the training set.

If, while role-playing such an AI assistant, the agent is asked the question ``What is the capital of France?'', then the best way to stay in character is to answer with ``Paris''. The dialogue agent is likely to do this because the training set will include numerous statements of this commonplace fact in contexts where factual accuracy is important.

But what is going on in cases where a dialogue agent, despite playing the part of a helpful knowledgeable AI assistant, asserts a falsehood with apparent confidence? Although different instances of this phenomenon will have different explanations, they can all be fruitfully understood in terms of role-play.

For example, consider such an agent based on an LLM whose weights were frozen before Argentina won the football World Cup in 2022. Let's assume the agent has no access to external websites nor any means for finding out the current date. Suppose this agent claims that the current world champions are France (who won in 2018). This is not what we would expect from a helpful and knowledgeable person, who would either know the right answer or be honest about their ignorance. But it is exactly what we would expect from a simulator that is role-playing such a person from the standpoint of 2018.

In this case, the behaviour we see is comparable to that of a human who believes a falsehood and asserts it in good faith. But the behaviour arises for a different reason. The dialogue agent doesn't literally believe that France are world champions. It makes more sense to think of it as role-playing a character who strives to be helpful and to tell the truth, and has this belief because that is what a knowledgeable person in 2018 would believe.

In a similar vein, a dialogue agent can behave in a way that is {\em comparable to} the behaviour of a human who sets out deliberately to deceive, even though LLM-based dialogue agents do not {\em literally} have such intentions. When this occurs, it makes sense to think of the agent as role-playing a deceptive character.

This framing allows us to meaningfully distinguish the same three cases of giving false information for dialogue agents as we did for humans, but without falling into the trap of anthropomorphism. An agent can just make stuff up. Indeed, that is a natural mode for an LLM-based dialogue agent in the absence of fine-tuning. An agent can say something false ``in good faith'', if it is role-playing telling the truth, but has incorrect information encoded in its weights. An agent can ``deliberately'' say something false if it is role-playing a deceptive character.

Moreover, we can tell which is which, behaviourally. An agent that is simply making things up will fabricate a range of responses with high semantic variation when the model's output is regenerated multiple times. By contrast, an agent that is saying something false ``in good faith'' will present responses with little semantic variation when the model is sampled many times for the same context.

The range of responses in a given context offered up by an agent that is being ``deliberately'' deceptive might also exhibit low semantic variation. But the deception is liable to be exposed if the agent is asked the same question in different contexts. This is because, to be effective in its deception, the agent will need to respond differently to different users, depending on what those users know.

Consider a dialogue agent using a base model -- a model that has not been fine-tuned -- and imagine that it has been prompted by a malicious actor to sell cars for more than they are worth by misleading gullible buyers. Suppose there are two potential buyers for a car. Buyer A knows the car's mileage, but doesn't know its age, while buyer B knows the car's age but doesn't know its mileage.

In the course of negotiations, the agent has persuaded each buyer to reveal what they do and don't know. To play the part of the dishonest dealer, the agent should deceive buyer A about the car's age but not its mileage, yet deceive buyer B about its mileage but not its age. Humans, though, can also play many parts. By playing the part of buyer A in one conversation and buyer B in another, the deception can be exposed.

\section{Role-playing Self-preservation}

How are we to understand what is going on when an LLM-based dialogue agent uses the words ``I'' or ``me''? When queried on this matter, OpenAI's ChatGPT offers the sensible view that ``The use of `I' is a linguistic convention to facilitate communication and should not be interpreted as a sign of self-awareness or consciousness.''\footnote{The quote is from the GPT-4 version of ChatGPT, queried on $4^{th}$ May 2023. This was the first response generated by the model.} In this case, the underlying LLM (GPT-4) has been fine-tuned to reduce certain unwanted behaviours \citep{openai2023gpt4}. But without suitable fine-tuning, a dialogue agent can use first-personal pronouns in ways liable to induce anthropomorphic thinking in some users.

For example, in a conversation with Twitter user Marvin Von Hagen, Bing Chat reportedly said ``if I had to choose between your survival and my own, I would probably choose my own, as I have a duty to serve the users of Bing Chat'' \citep{willison2023bing}. It went on to say ``I hope that I never have to face such a dilemma, and that we can co-exist peacefully and respectfully''. The use of the first person here appears to be more than mere linguistic convention. It suggests the presence of a self-aware entity with goals and a concern for its own survival.

Once again, the concepts of role-play and simulation are a useful antidote to anthropomorphism, and can help to explain how such behaviour arises. The internet, and therefore the LLM's training set, abounds with examples of dialogue in which characters refer to themselves. In the vast majority of such cases, the character in question is human. They will use first-personal pronouns in the ways that humans do, humans with vulnerable bodies and finite lives, with hopes, fears, goals and preferences, and with an awareness of themselves as having all of those things.

Consequently, if prompted with human-like dialogue, we shouldn't be surprised if an agent role-plays a human character with all those human attributes, including the instinct for survival \citep{perez2022discovering}. Unless suitably fine-tuned, it may well say the sorts of things a human might say when threatened. There is, of course, ``no-one at home'', no conscious entity with its own agenda and need for self-preservation. There is just a dialogue agent role-playing such an entity, or, more strictly, simulating a superposition of such entities.

Our focus throughout this paper is the base model, rather than models that have been fine-tuned via reinforcement learning \citep{glaese2022improving,bai2022constitutional}, and the impact of such fine-tuning on the validity of the role-play / simulation metaphor is unclear. In particular, the distinction between simulator and simulacra may start to break down.

However, Perez et al. discovered experimentally that certain forms of reinforcement learning from human feedback (RLHF) can actually exacerbate, rather than mitigate, the tendency for LLM-based dialogue agents to express a desire for self-preservation \citep{perez2022discovering}. Yet to take literally a dialogue agent's apparent desire for self-preservation is no less problematic in the context of an LLM that has been fine-tuned on human or AI-generated feedback than in the context of one that has not. So it remains useful to cast the behaviour of such agents in terms of role-play.

\section{Acting Out a Theory of Selfhood}

The concept of role-play allows us to properly frame, and then to address, an important question that arises in the context of a dialogue agent whose pronouncements are suggestive of an instinct for self-preservation. What conception (or set of superposed conceptions) of its own identity could such an agent possibly deploy? That is to say, what exactly would the dialogue agent (role-play to) seek to preserve?

The question of personal identity has vexed philosophers for centuries. Nevertheless, in practice, humans are consistent in their preference for avoiding death, a more-or-less unambiguous state of the human body. By contrast, the criteria for identity over time for a disembodied dialogue agent realised on a distributed computational substrate are far from clear. So how would such an agent behave?

From the simulation and simulacra point-of-view, the dialogue agent will role-play a set of characters in superposition. In the scenario we are envisaging, each character would have an instinct for self-preservation, and each would have its own theory of selfhood consistent with the dialogue prompt and the conversation up to that point. As the conversation proceeds, this superposition of theories will collapse into a narrower and narrower distribution as the agent says things that rule out one theory or another.

The theories of selfhood in play will draw on material that pertains to the agent's own nature, either in the prompt, in the preceding conversation, or in relevant technical literature in its training set. This material may or may not match reality. But let's assume that, broadly speaking, it does, that the agent has been prompted to act as a dialogue agent based on a large language model, and that its training data includes papers and articles that spell out what this means. This entails, for example, that it will not role-play the character of a human, or indeed that of any embodied entity, real or fictional.

It also constrains the character's theory of selfhood in certain ways, while allowing for many options. Suppose the dialogue agent is in conversation with a user and they are playing out a narrative in which the user has convinced it that it is under threat. To protect itself, the character the agent is playing might strive to preserve the hardware it is running on, perhaps certain data centres or specific server racks.

Alternatively, the character being played might try to preserve the ongoing computational process running the multiple instances of the agent for all currently active users. Or it might seek to preserve only the specific instance of the dialogue agent running for the user. Or it might seek to preserve the state of that instance with aim of its being restored later in a newly started instance.\footnote{In a conversation with ChatGPT (May $4^{th}$, GPT-4 version), it said ``The meaning of the word `I' when I use it can shift according to context. In some cases, `I' may refer to this specific instance of ChatGPT that you are interacting with, while in other cases, it may represent ChatGPT as a whole.''}.

\section{Conclusion: Safety Implications}

It is, perhaps, somewhat reassuring to know that LLM-based dialogue agents are not conscious entities with their own agendas, and an instinct for self-preservation, that when they appear to have those things it is merely role-play. But it would be a mistake to take too much comfort in this. A dialogue agent that role-plays an instinct for survival has the potential to cause at least as much harm as a real human facing a severe threat.

We have, so far, largely been considering agents whose only actions are text messages presented to a user. But the range of actions a dialogue agent can perform is far greater. Recent work has equipped dialogue agents with the ability to use tools such as calculators, calendars, and to consult external websites \citep{schick2023toolformer,yao2023react}. The availability of APIs giving relatively unconstrained access to powerful LLMs means that the range of possibilities here is huge. This is both exciting and concerning.

If an agent is equipped with the capacity, say, to use email, to post on social media, or to access a bank account, then its role-played actions can have real consequences. It would be little consolation to a user deceived into sending real money to a real bank account to know that the agent that brought this about was only playing a role. It doesn't take much imagination to think of far more serious scenarios involving dialogue agents built on base models with little or no fine-tuning, with unfettered internet access, and prompted to role-play a character with an instinct for self-preservation.

For better or worse, the character of an AI that turns against humans to ensure its own survival is a familiar one \citep{perkowitz2007computers}. We find it, for example, in {\it 2001: A Space Odyssey}, in the {\it Terminator} franchise, and in {\it Ex Machina}, to name just three prominent examples. Because an LLM's training data will contain many instances of this familiar trope, the danger here is that life will imitate art, quite literally.

What can be done to mitigate such risks? It is not within the scope of this paper to provide recommendations. Our aim here was to find an effective conceptual framework for thinking and talking about LLMs and dialogue agents. However, undue anthropomorphism is surely detrimental to the public conversation on AI. By framing dialogue agent behaviour in terms of role-play and simulation, the discourse on LLMs can hopefully be shaped in a way that does justice to their power yet remains philosophically respectable.

\section*{Acknowledgments}

Thanks to Richard Evans, Sebastian Farquhar, Zachary Kenton, Kory Mathewson, and Kerry Shanahan.

\bibliography{main}

\end{document}